\documentclass{article}

\usepackage{microtype}
\usepackage{graphicx}
\usepackage{subfigure}
\usepackage{booktabs} 

\usepackage{hyperref}

\usepackage[accepted]{icml2023}

\usepackage{amsmath}
\usepackage{amssymb}
\usepackage{mathtools}
\usepackage{amsthm}
\usepackage{array}
\usepackage{comment}

\usepackage[capitalize,noabbrev]{cleveref}

\usepackage{xcolor}
\definecolor{airforceblue}{rgb}{0.14, 0.31, 0.5}
\definecolor{brightmaroon}{rgb}{0.76, 0.13, 0.28}
\definecolor{mdgreen}{rgb}{0.05,0.6,0.05}
\definecolor{mdairforceblue}{rgb}{0.1, 0.3, 0.7}
\definecolor{mdblue}{rgb}{0,0,0.7}
 \newcommand{\blue}[1]{ \textcolor{mdairforceblue}{#1}}

\newcommand{\green}[1]{ \textcolor{mdgreen}{#1}}

\definecolor{orange}{rgb}{1,0.5,0}
\definecolor{dkblue}{rgb}{0,0,0.5}
\definecolor{dkgray}{rgb}{0.3,0.3,0.3}
\definecolor{slate}{rgb}{0.25,0.25,0.4}
\definecolor{gray}{rgb}{0.5,0.5,0.5}
\definecolor{ltgray}{rgb}{0.7,0.7,0.7}
\definecolor{purple}{rgb}{0.7,0,1.0}
\definecolor{lavender}{rgb}{0.65,0.55,1.0}

\newcommand{\ensuretext}[1]{#1}

\newcommand{\marker}[2]{\ensuremath{^{\textsc{#1}}_{\textsc{#2}}}}
\newcommand{\authcomment}[3]{\ensuretext{\textcolor{#3}{[#1 #2]}}}

\newcommand{\tushar}[1]{\authcomment{\marker{T}{K}}{#1}{orange}}

\newcommand{\ashish}[1]{\authcomment{\marker{A}{S}}{#1}{mdgreen}}
\newcommand{\resolved}[1]{}

\theoremstyle{plain}

\theoremstyle{definition}

\theoremstyle{remark}

\usepackage[textsize=tiny]{todonotes}

\icmltitlerunning{Specializing Smaller Language Models towards Multi-Step Reasoning}

\begin{document}

\twocolumn[
\icmltitle{Specializing Smaller Language Models towards Multi-Step Reasoning 
           }



\icmlsetsymbol{equal}{*}

\begin{icmlauthorlist}
\icmlauthor{Yao Fu$^{\spadesuit}$}{}
\icmlauthor{Hao Peng$^{\clubsuit}$}{}
\icmlauthor{Litu Ou$^{\spadesuit}$}{}
\icmlauthor{Ashish Sabharwal$^{\clubsuit}$}{}
\icmlauthor{Tushar Khot$^{\clubsuit}$}{}
\end{icmlauthorlist}
\center{$^{\spadesuit}$University of Edinburgh\quad\quad $^{\clubsuit}$Allen Institute for AI}
\center{\{yao.fu, s1970716\}@ed.ac.uk\quad\quad\{haop, ashishs, tushark\}@allenai.org}


\icmlkeywords{Machine Learning, ICML}

\vskip 0.3in
]

\begin{abstract}
The surprising ability of Large Language Models (LLMs) to perform well on complex reasoning with only few-shot chain-of-thought prompts is believed to emerge only in very large-scale models (100+ billion parameters). 
We show that such abilities can, in fact, be distilled down from GPT-3.5 ($\ge$ 175B) to T5 variants ($\le$ 11B). 
We propose \textit{model specialization}, to specialize the model's ability towards a target task. 
The hypothesis is that 
large models (commonly viewed as larger than 100B) have strong modeling power, but are spread on a large spectrum of tasks. 
Small models (commonly viewed as smaller than 10B) have limited model capacity, but if we concentrate their capacity on a specific target task,
the model can achieve a decent improved performance.
We use multi-step math reasoning as our testbed because it is a very typical emergent ability.
We show two important aspects of model abilities:
(1). there exists a very complex balance/ tradeoff between language models' multi-dimensional abilities;
(2). by paying the price of decreased generic ability, we can clearly lift up the scaling curve of models smaller than 10B towards a specialized multi-step math reasoning ability.
We further give comprehensive discussions about important design choices for better generalization, including the tuning data format, the start model checkpoint, and a new model selection method. 
We hope our practice and
discoveries can serve as an important attempt towards specialized smaller models in the new research paradigm set by LLMs.
\end{abstract}

\section{Introduction}
\label{sec:intro}

Recently, the field of NLP is significantly impressed by large language models' strong abilities~\citep{brown2020language, chowdhery2022palm}. 
\citet{wei2022emergent} discuss the emergent abilities of large language models -- abilities that seems to only exist in large models (more than 100B parameters), but not in small models. 
A very typical example (also the first discovered emergent ability) is to perform multi-step reasoning on math word problems by chain-of-thought (CoT) prompting~\citep{wei2022chain} where the authors let the model generate a step-by-step reasoning chain to help get the final answer. 
The existence of such abilities has a very deep, profound influence on the community: on the positive side, such abilities open countless opportunities for new research directions; on the negative side, very few organizations have the compute to even fine-tune 100B-scale models, making the accessibility of such abilities extremely hard. 
It would be ideal if smaller models can also obtain emergent abilities like math CoT reasoning, so they can be accessed by a larger range of researchers and practitioners. 
However, preliminary results of \citet{wei2022emergent} show that if the model scale is small (empirically less than 100B parameters), CoT exhibits flat, sometimes even near zero scaling curve~\citep{wei2022chain}. 
Later smaller models' scaling curve is partially improved in~\citet{chung2022scaling}, but still worse than large models. 
These results so far are rather pessimistic since they suggest increasing CoT performance for smaller models can be challenging. 
At the current stage, the community is eager to know to what extent such abilities can be further improved in smaller models.

This paper addresses the problem of CoT reasoning for smaller models by \textit{model specialization}.
Our hypothesis is that large models ($\ge$ 100B) have strong modeling power but are spread over a large spectrum of tasks.
Small models ($\le$ 10B) have limited model capacity, but if we concentrate their capacity on a target task, the model may still have a decent improved performance.
There exists promising preliminary work on smaller models' chain-of-thought abilities such as UL2~\citep{tay2022unifying} and FlanT5~\citep{chung2022scaling}, but they focus on generic abilities and consequently, the model's (limited) power is not concentrated. 
In our experiments, we show that by paying the price of decreased abilities in generic tasks (specifically we lose a large portion of  accuracy on the BigBench Hard suite~\citealp{suzgun2022challenging}), we can lift the scaling curve of CoT reasoning on small FlanT5 models (250M, 760M, and 3B) by a large margin (an average +10 accuracy gain) on a suite of 4 math reasoning tasks (1 in-distribution and 3 out-of-distribution). 
This means that we can indeed move the model's power from generic abilities to concentrate on the target math CoT.

Our approach is to fine-tune an instruction-tuned model (FlanT5) by distilling chain-of-thought reasoning paths of the GSM8K data from a large teacher model (GPT-3.5 code-davinci-002~\citealp{chen2021evaluating}), then do a model selection on the average performance of three held-out math reasoning data to ensure the model's out-of-distribution generalization.  
Although distillation per se is a well-studied area,
there are multiple caveats in our process, as we will demonstrate:
(1). the teacher model code-davinci-002 and our student model FlanT5 use different tokenizers, we address the tokenizer alignment problem by dynamic programming.
(2). Distillation induces different performance on an instruction-tuned checkpoint (in our case, FlanT5) and the raw pretrained checkpoint (T5), where specialized FlanT5 performs better but specialized T5 achieves more accuracy gain.
(3). at the late training stage, the model's in-distribution and out-of-distribution (OOD) performance fluctuates differently, so if one wants better OOD generalization, the model selection should be performed on held-out math datasets, rather than the validation portion of the tuning data.
(4). multiple tradeoffs happen during the distillation/ specialization process: as we start distillation, on BigBench Hard test suite (the measure of generic ability), the model immediately loses 
all its CoT prompting abilities, and gradually loses a large portion (but not all) of answer-only prompting abilities. 
The data format we use for tuning is also closely related to model ability: in-context examples enable both in-context and zero-shot performance,
but zero-shot examples lose the model's in-context ability for increased zero-shot ability.

These findings deepen our understanding of language model chain-of-thought reasoning behavior in multiple aspects:
(1). the previous hypothesis is that CoT has near-flat scaling curves on small scale, we show that we can lift up the scaling curve by concentrating the model's capacity on a target ability.
This indicates that chain-of-thought might not be an emergent ability because, after specialization, smaller models' scaling curves become log-linear, just like large models~\citep{kaplan2020scaling, hoffmann2022training}. 
(2). previous observation of LLM behaviors indicates complex tradeoffs and balances of model ability across multiple dimensions, we give a detailed description of how we move the model's power from generic abilities to a target ability, clearly showing what can be gained at what cost. 
(3). classical model selection theory selects the model on the validation portion of the same dataset, we select the model based on the performance of different math reasoning datasets, to prevent overfitting on one single dataset. 
We hope our practice and discoveries can serve as an example 
attempt
towards strong specialized smaller models.

\section{Background}
\label{sec:background}
\begin{figure*}[!t]
\small
  \centering
  \includegraphics[width=\linewidth]{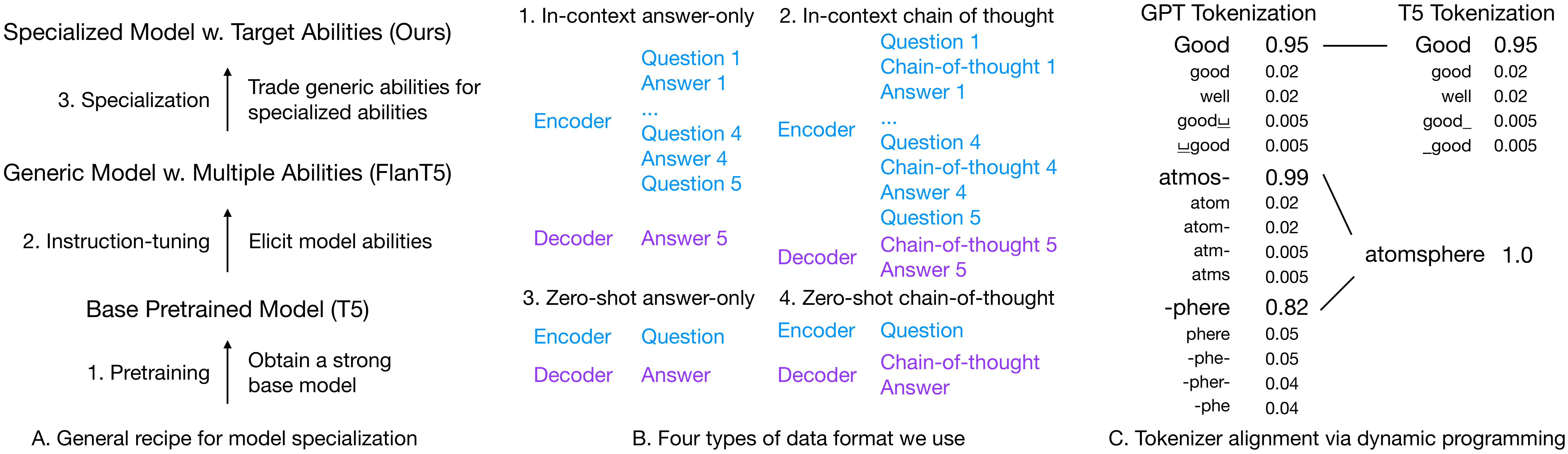}
  \caption{\textbf{A.} Model specialization process. Pretraining gives a strong base model~\citep{raffel2020exploring, chowdhery2022palm}, instruction tuning elicits the model ability~\citep{chung2022scaling}, then specialization (this work's focus) moves model abilities to a target direction. In this work, we trade the model's generic abilities (as measured by BigBench Hard) for the model's multi-step math reasoning abilities. \textbf{B.} Four data formats we consider for tuning the model. 
  We will show tuning with in-context chain-of-thought examples is particularly important for the model's CoT ability. \textbf{C.} Aligning GPT tokenization to T5 tokenization by dynamic programming. If a T5 token has a one-to-one alignment to a GPT token, we reuse the GPT's top 5 probability as the target distribution. If there the mapping is one-to-many/ many-to-one, we treat the T5 token's distribution as one-hot. 
  }
  \label{fig:method}
\end{figure*}

\textbf{Large Language Models' Abilities}\quad\quad
Large language models have significantly changed the research paradigm in NLP by showing strong abilities on multiple dimensions~\citep{brown2020language, hoffmann2022training, chowdhery2022palm, wei2022emergent}.
Currently, the new recipe for training LLMs is to first train a base model (e.g., GPT-3, PaLM, OPT), then elicit the abilities of the base model by instruction tuning (e.g., GPT-3 $\to$ InstructGPT~\citealp{ouyang2022training}; PaLM $\to$ FlanPaLM~\citealp{chung2022scaling}, OPT $\to$ OPT-IML~\citealp{iyer2022opt}, also see Fig.~\ref{fig:method}A step 1 and 2).
For the base model, initially,~\citet{wei2022chain} shows that the chain-of-thought performance curve is near-zero if the model size is smaller than 100B. 
Later~\citet{chung2022scaling} updated this hypothesis by showing CoT can be unlocked if CoT data is included as one particular type of instruction, 
but their model's performance is not as good because their model's ability is spread over multiple dimensions.
This work shows that CoT performance can be significantly lifted if we concentrate model's power toward a target ability (Fig.~\ref{fig:method}A, step 3). 


\textbf{Specialized Language Models} \quad\quad
Although modern language models show strong generic abilities on multiple directions, recent analysis~\citep{fu2022gptroadmap} shows models do have different focuses (e.g., code-davinci-002 for code and text-davinci-003 for text). 
Ability tradeoff happens at all scale: 
for large models, such a tradeoff does not have to be all or nothing: code-davinci-002, although specialized for code, can still solve a lot of text problems;
for small models, due to limited model capacity, they have to trade all generic abilities for one special ability.
One example is GitHub Copilot, which supposedly is a 12B small model~\citep{copilotexplorer}. 
The actual practice of specialization is simply finetuning: to specialize a model towards a target ability, one simply tunes the model using the related data, which is the practice of concurrent work about smaller models' CoT ability~\citep{magister2022teaching, shridhar2022distilling, ho2022large}. 
The problem here is how to generalize beyond the tuning data, as small models may simply overfit the tuning distribution but struggle to generalize when the distribution shifts~\citep{liu-etal-2022-challenges, si2022prompting}.
So far the community's hypothesis of OOD generation involves two important aspects: (1). model scale~\citep{chowdhery2022palm}; (2). instruction tuning~\citep{chung2022scaling}, which we will also study. 
These factors mark the differences between our work and the concurrent distillation work: we show how the model trades generic abilities for the target ability, and how model scale and instruction tuning help the model gain better in-distribution and OOD performance. 
%


\textbf{Distillation and Data Augmentation}\quad\quad 
Our approach of using data generated from code-davinci-002 to tune smaller FlanT5 can be viewed as either distillation~\citep{tan2018multilingual} or data augmentation~\citep{li2022explanations}. 
Here we note that we merely use the generated data as the tool for model specialization, and the specialization data can also be from other sources like human annotation. 
Our focus is to study the ability tradeoff during specialization, but not directly contribute to the distillation or data augmentation literature. 


\textbf{Most closely related works}\quad\quad There are two threads of most related works: (1). FlanT5~\citep{chung2022scaling} and UL2~\citep{tay2022unifying} which is the first work discussing smaller models' CoT ability, but they focus on generic CoT while we trade generic ability for math CoT. 
(2). language model self-improvement~\citep{huang2022large} which also use CoT data augmentation, but they only consider large models and do not show the tradeoff between model abilities. Here we focus on small models and clearly show the price for ability improvements. 


\section{Specializing Multi-Step Reasoning}
\label{sec:method}
Our objective is to study what it takes to improve smaller models' chain-of-thought math reasoning. 
We use GSM8K~\citep{cobbe2021training} as our seed dataset because it is one of the  datasets with most diverse math reasoning problems,
but test the model's performance of three additional math datasets (MultiArith, ASDiv, and SVAMP~\citealp{wei2022chain}) to show the model generalizes to OOD data. 
We further use BigBench Hard to test to model's generic reasoning ability, demonstrating the tradeoff between generic and target abilities. 
We use T5 (raw pretrained checkpoint) and FlanT5 (instruction tuned checkpoint) as our base model, and use code-davinci-002 to generate distillation/ specialization data. 

\textbf{Distillation from Code-Davinci-002}\quad\quad
Given a training question corpora,
we use code-davinci-002 to generate 40 new CoT solutions then take the 
ones that lead to the correct answers as our training data.
One solution consists of an answer and a chain of thought explaining the intermediate steps towards the answer. 
In addition to the standard finetuning setting where one uses the question as the input and use the [CoT, answer] pair as the output (Fig.~\ref{fig:method} B4), we further consider three additional data formats:
(1). in-context answer-only (Fig.~\ref{fig:method} B1), where we do not use the CoT data (hence the name ``answer-only'') and prepend 4 in-context examples before the question (hence the name ``in-context''). 
The reason we prepend the in-context example is that previous work shows tuning with in-context examples improves the model's in-context learning ability~\citep{min-etal-2022-metaicl}. 
(2). in-context chain-of-thought (Fig.~\ref{fig:method} B2), where we add CoT to both the in-context example and the output. 
(3). zero-shot answer-only, where we directly input the question and output the answer. 
Using answer-only data is because previous work shows they improve performance. 
In our experiments, we will show that in-context data induces zero-shot ability but zero-shot data sacrifice in-context learning ability. 
We note that there also exist techniques like adding a calculator~\citep{cobbe2021training} or self-consistency decoding~\citep{wang2022self} that can further improve the performance.
These techniques are orthogonal to the distillation we use and can definitely be integrated to our work for better performance.
Since our focus is the balance of the models' special and generic abilities, we leave the integration of these orthogonal techniques to future work.

In terms of training objectives, in the distillation literature, there are typically two types of distillation approaches: 
(1). sample matching, where one trains the student model on the data generated by the teacher. 
In our case, sample matching means we directly optimize the student's likelihood on the data generated by code-davinci-002. 
(2). distribution matching, where one minimizes the KL divergence between the student's output distribution 
(in our case, the per-step autoregressive distribution) and the teacher's. 
Usually, distribution matching is shown to achieve faster convergence and better performance than sample matching, so we use distribution matching as our training objective. 
Distribution matching has an additional challenge in storing the distribution parameter: at each step, we need to store the whole distribution defined on the vocabulary $\mathcal{V}$, so the size of the dataset is $|\mathcal{V}|$ times larger than sample matching.
Yet the OpenAI API only grants access to the 5 most probable tokens at each decoding step, but not the probability distribution over the entire vocabulary. 
Although the per-step distribution only covers the top 5 tokens, most of the time their probability sum is close to 1, being a good enough approximation of the full vocabulary distribution. 
We set to zero the probabilities of tokens not in the top 5.


\textbf{Aligning tokenizers by dynamic programming}\quad\quad One problem when matching the two distributions is the misalignment between the GPT tokenizer and the T5 tokenizer.
We solve this problem by dynamic programming. 
Specifically, given two sequences to tokens $[\mathbf{s}_{1:L}, \mathbf{t}_{1:N}]$ ,
our objective is to find an alignment  that minimizes the total cost of editing one sequence to the other. 
Our dynamic program is a slight tweak of the textbook dynamic programming algorithms used in 
bioinformatics for sequence alignment (such as the Needleman–Wunsch algorithm~\cite{Needleman1970AGM}) and in signal processing (such as dynamic time wrapping~\cite{Senin2008DynamicTW}).  
The recursion function is: 
\begin{align}
    f(i, j) = \min\{&f(i-1, j) + c(\mathbf{s}_{i}, \mathbf{t}_{j}), \label{eq:1}\\
    &f(i, j-1) + c(\mathbf{s}_i, \mathbf{t}_{j}), \label{eq:2}\\
    &f(i - 1, j - 1) + c(\mathbf{s}_i, \mathbf{t}_j)\} \label{eq:3}
\end{align}
where $f(i, j)$ denotes the total cost aligning $\mathbf{s}_{1:i}$ and $\mathbf{t}_{1:j}$ and $c(\mathbf{s}_{i}, \mathbf{t}_{j})$ is the predefined string edit distance between token $\mathbf{s}_i$ and $\mathbf{t}_j$. 
our algorithm does not enforce one-on-one matching between tokens in the two sequences, and one token in $\mathbf{s}$ might align with multiple in $\mathbf{t}$ and vice versa
Fig.~\ref{fig:method}C gives an example alignment. 
If there exists a one-to-one mapping between a GPT token and a T5 token, we use the GPT distribution as the T5 distribution.
If the mapping is not one-to-one, e.g., two T5 tokens map to one GPT token, or two GPT tokens map to one T5 token (Fig.~\ref{fig:method} C lower part), we do not use the corresponding GPT distribution and set the T5 distribution to be one-hot. 
We further note that aligning sequences generated by different tokenizers is a generic problem of contemporary NLP, yet we are not aware of any existing libraries approaching it. 
We plan to release the implementation of our dynamic program and hope it can be useful for future research.




\section{Experiments}
\label{sec:experiments}

\newcommand{\sepsmall}[0]{\hspace{4pt}}
\newcommand{\septiny}[0]{\hspace{8pt}}
\begin{table*}[th!]
  \caption{
  Overall test set performance. We specialize Flan-T5's ability from the generic tasks (BigBench Hard) to math reasoning tasks.
  After paying the cost of BigBench Hard performance (the model loses all the CoT prompting ability and a large portion of the Answer-only (AO) prompting ability), we see the specialized T5 models have improved in-distribution (GSM8K) performance (where our 3B and 11B models outperform concurrent works) as well as out-of-distribution (MultiArith, ASDiv and SVAMP) performance, showing that we can move the model's ability from generic tasks (BBH) to a specific target task (math reasoning). 
  Magister22: \citet{magister2022teaching};
  Shridhar22: \citet{shridhar2022distilling};
  Ho22: \citet{ho2022large}.
  }
  \label{tab:exp:overall}
  \centering
  \setlength{\tabcolsep}{4pt}
  \begin{tabular}{
            @{}l @{} r  @{\hskip 16pt}
            c@{\sepsmall}c  m{0.01em} 
            c@{\sepsmall}c  m{0.01em} 
            c@{\sepsmall}c  m{0.01em} 
            c@{\sepsmall}c  m{0.01em} 
            @{\hskip 8pt}
            c@{\sepsmall}c  m{0.01em} 
            c@{\sepsmall}c 
            @{}}
      \toprule[1.25pt]
      
       &&  \multicolumn{11}{c}{\textbf{CoT Reasoning on Maths Word Problems}} && \multicolumn{5}{c}{\textbf{BigBench-Hard}} \\ 
       \cmidrule(lr){3-13}
       \cmidrule(lr){15-19}
        & & \multicolumn{2}{c}{\textbf{GSM8K}} && \multicolumn{2}{c}{\textbf{MultiArith}} && \multicolumn{2}{c}{\textbf{ASDiv}} && \multicolumn{2}{c}{\textbf{SVAMP}} 
        && \multicolumn{2}{c}{\textbf{AO}} && \multicolumn{2}{c}{\textbf{CoT}} \\
       \cmidrule(lr){3-4}
       \cmidrule(lr){6-7}
       \cmidrule(lr){9-10}
       \cmidrule(lr){12-13}
       \cmidrule(lr){15-16}
       \cmidrule(lr){18-19}
        \textbf{Models}
        & \textbf{\#Params.} 
        &   \textbf{Acc.} & $\mathbf{\Delta}$
        &&  \textbf{Acc.} & $\mathbf{\Delta}$
        &&  \textbf{Acc.} & $\mathbf{\Delta}$
        &&  \textbf{Acc.} & $\mathbf{\Delta}$
        &&  \textbf{Acc.} & $\mathbf{\Delta}$
        &&  \textbf{Acc.} & $\mathbf{\Delta}$\\
        
      \midrule[1.25pt]
        
        code-davinci-002 & $\ge$175B 
        &  63.1 & - 
        && 95.8 & - 
        && 80.4 & - 
        && 76.4 & - 
        && 56.6 & - 
        && 73.9 & -\\ 

        LaMDA & 137B 
        &  14.8 & - 
        && 45.0 & - 
        && 46.6 & - 
        && 37.5 & - 
        && -    & - 
        && -    & -  \\ 

        PaLM & 60B 
        &  29.9 & - 
        && 75.0 & - 
        && 61.9 & - 
        && 46.7 & - 
        && 37.4 & - 
        && 43.0 & - \\ 

        UL2 & 20B 
        & \phantom{0}4.4 & - 
        && -    & - 
        && 16.9 & - 
        && 12.5 & - 
        && -    & - 
        && -    & - \\ 

      \midrule[1.25pt]
        \multicolumn{18}{@{}l}{\textbf{Concurrent Works with Knowledge Distillation}}\\
      Magister22, T5
      & 11B 
      &  21.9  & - 
      && -    & - 
      && 42.1 & - 
      && -    & - 
      && ?    & - 
      && ?    & - \\

      Shridhar22, GPT
      & 6B 
      & 21.0 & -  
      && -   & -  
      && -   & - 
      && -   & - 
      && ?   & - 
      && ?   & - \\

      Ho22, GPT
      & 6B 
      & \phantom{0}6.8 & - 
      && 33.3 & - 
      && -    & - 
      && -    & - 
      && ?    & - 
      && ?    & - \\ 

      \midrule[1.25pt]
      \multicolumn{18}{@{}l}{\textbf{Our Specialized Models Compared with Baselines}}\\
      
      FlanT5-XXL 
      & 11B 
      &  16.1 & - 
      && 51.7 & - 
      && 36.5 & - 
      && 39.7 & - 
      && 47.4 & - 
      && 41.8 & - \\

      \; + Specialized 
      &  11B  
      &  27.1  & \green{+11.0} 
      && 63.0 & \green{+11.3} 
      && 37.6 & \phantom{0}\green{+1.1} 
      && 35.6 & \phantom{0}\blue{-4.1} 
      && 19.6 & \blue{-27.8} 
      && \phantom{0}0.0  & \blue{-41.8} \\

      \midrule[0.5pt]

      FlanT5-XL 
      & 3B 
      &  13.5 & -
      && 24.0   & -  
      && 20.7 & - 
      && 17.7 & - 
      && 39.9 & - 
      && 35.8 & - \\ 
      \; + Specialized 
      & 3B  
      &  22.4 & \phantom{0}\green{+8.9} 
      && 42.3 & \green{+18.3} 
      && 28.4 & \phantom{0}\green{+7.7} 
      && 23.8 & \phantom{0}\green{+6.1} 
      && \phantom{0}3.2  & \blue{-36.7} 
      && \phantom{0}0.0  & \blue{-35.8} \\ 
      
      \midrule[0.5pt]
      
      FlanT5-Large 
      & 760M 
      &  \phantom{0}6.9  & - 
      && 13.0 & - 
      && 10.1 & - 
      && \phantom{0}6.8  & - 
      && 30.3 & - 
      && 30.9 & - \\

      \; + Specialized 
      & 760M  
      &  20.2 & \green{+13.3} 
      && 38.5 & \green{+25.5} 
      && 23.8 & \green{+13.7} 
      && 20.4 & \green{+13.6} 
      && \phantom{0}6.5  & \blue{-23.8} 
      && \phantom{0}0.3 & \blue{-30.6} \\ 
      
      \midrule[0.5pt]
      
      FlanT5-Base 
      & 250M 
      &  \phantom{0}3.0  & - 
      && \phantom{0}7.0  & - 
      && \phantom{0}4.2  & - 
      && \phantom{0}3.8  & - 
      && 24.2 & - 
      && 25.9 & - \\ 
      
      \; + Specialized
      & 250M  
      &  13.4 & \green{+10.4} 
      && 29.7 & \green{+22.7} 
      && 20.9 & \green{+16.7} 
      && 14.2 & \green{+10.4} 
      && \phantom{0}3.1  & \blue{-21.1} 
      && \phantom{0}0.1 & \blue{-25.8}\\ 
      
      \bottomrule[1.25pt]
  \end{tabular}
\end{table*}

The objective of the experiments is to see to what extent we can lift up the scaling curve of smaller models' math CoT performance and what is the price of it. 
We conduct model specialization on two model families: the raw pretrained checkpoints, and their instruction-tuned checkpoints (recall that the instruction-tuned checkpoints are generally more capable than the raw pretrained checkpoints, Fig~\ref{fig:method}A). 
Specifically, we consider the raw pretrained T5 Base (250M)/ Large (760M)/ XL (3B)/ XXL (11B), and the instruction-tuned FlanT5s. 
In Sec.~\ref{ssec:exp:overall}, we validate our main hypothesis that large models can perform well on a wide range of tasks while smaller model's ability can be moved from generic abilities to a specialized target ability.
Specifically, we show model specialization can indeed improve CoT math performance for FlanT5-Base/ Large/ XL/ XXL, while paying the price of generic abilities, i.e., losing all CoT abilities on BigBench Hard and a large portion of answer-only (AO) abilities.
In Sec.~\ref{ssec:exp:scaling_curve}, we study the scaling behavior of smaller models and show how specialization lifts up the scaling curve for both T5 and FlanT5. 
This modifies the previous belief that smaller models exhibit a flat scaling curve~\citep{wei2022chain}; we show that their scaling curve becomes log-linear after specialization, but not flat. 
In Sec~\ref{ssec:exp:dynamics}, we show the dynamics and the generalization behavior of specialization: the model's target performance increases gradually but generic abilities decrease gradually during tuning, and there exists tradeoffs between in-distribution v.s. OOD performance and in-context v.s. zero-shot performance. 


\subsection{Overall Performance Tradeoff}
\label{ssec:exp:overall}
We test the models' math reasoning ability and generic ability and show their tradeoffs. 
For the math reasoning ability, 
we use the code-davinci-002 augmented GSM8K dataset~\citep{cobbe2021training} as our tuning dataset.
The GSM8K has 7K training questions, for each question we ask the large model to generate 40 different solutions, taking the correct ones from the generation, we have 130K tuning data points in total. 
We test the model's out-of-distribution performance on MultiArith, ASDiv, and SVAMP (collectively denoted as M-A-S) datasets~\citep{wei2022chain}.
None of the datasets has official train-dev-test splits, so we randomly sample 500 instances as the validation set,
and use the remaining instances (800 for GSM8K, 400 for MultiArith, 18K for ASDiv, 500 for SVAMP) as the test set.
The difference between M-A-S and GSM8K is that they are all primary school level arithmetic reasoning problems, but the entities involved in the datasets are different.
For example, GSM8K may consider arithmetic reasoning on foods (e.g, 5 apples + 8 bananas = 13 fruits) and MultiArith may consider animals (e.g., 2 dogs + 3 cats = 5 animals). 
This type of out-of-distribution generalization is usually referred to as lexical-level compositional generalization (i.e., both are addition, but the lexicons are different, see~\citealp{liu-etal-2022-challenges}). 
For the generic ability, we use BigBench Hard (BBH, ~\citealp{suzgun2022challenging}) test suite, a list of 26 challenging dataset testing the model's reasoning abilities from multiple dimensions (e.g., date understanding, causal judgement, referential game, .etc). 
Because of its difficulty and wide-coverage, BBH makes an ideal benchmark testing models' generic ability. 

For the baseline models, we consider generic large models and concurrent smaller distilled models, specifically:
(1). generic large models, ranked according to scale: code-davinci-002 (our teacher model, presumably larger or equal to 175B); LaMDA 137B~\citep{thoppilan2022lamda} and PaLM 60B~\citep{chowdhery2022palm}, both are strong generic models for chain-of-thought reasoning; UL2~\citep{tay2022unifying}, a 20B model with good CoT ability. We will show that specialized FlanT5 11B outperforms UL2 20B and becomes close to PaLM 60B and LaMDA 137B on the target math reasoning task. 
(2). concurrent works with knowledge distillation from~\citet{magister2022teaching, shridhar2022distilling, ho2022large}. We will show that our specialized FlanT5 clearly outperform all of them on the distillation data (with the cost of BBH performance), mostly because we use an instruction-tuned checkpoint (FlanT5) as the base model rather than the raw pretrained checkpoint (T5).

\begin{figure*}[!t]
\small
  \centering
  \includegraphics[width=\linewidth]{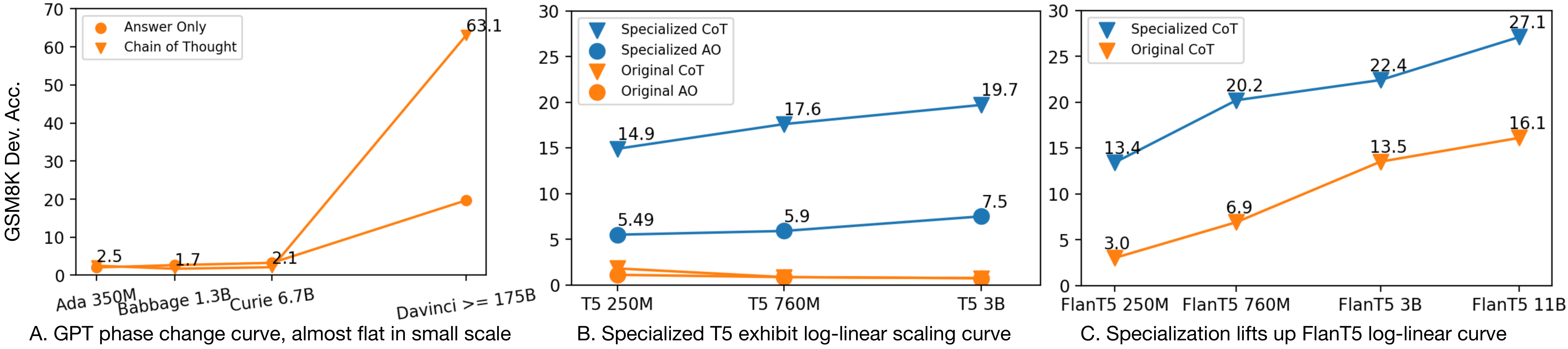}
  \caption{X-axis means log of model scale, y-axis means validation accuracy on GSM8K. 
  \textbf{A}: Previously, the community believe that small models has flat curve for both AO and CoT prompting and only when models become large enough the performance will have a ``phase change'' and suddenly increase. 
  \textbf{B}: we show that after training on CoT, the model exhibits log-linear curves where 
  both AO and CoT increase with model scale. 
  \textbf{C}: for instruction-tuned models (FlanT5) that already exhibit CoT, specialization lifts up the scaling curve, and the two curves are again, log-linear shaped. 
  All the log-linear curves indicate that chain-of-thought may not be an emergent ability which is marked by the flat-then-phase-change curve.
  Here we show the curve in small scale is not flat but actually log-linear, and continuously increasing model scale leads to continuously increased accuracy (no sudden phase change). 
  }
  \label{fig:exp:scaling}
\end{figure*}

\textbf{Trading generic abilities for math CoT reasoning}\quad\quad The overall results are in Table~\ref{tab:exp:overall}.
After tuning on the seed GSM8K augmented data, all FlanT5 models have improved math reasoning performance with approximately +10 average accuracy gain. 
We note that our smaller 3B model outperforms the current 11B and 6B distillation models on the GSM8K test set.
Despite multiple confounders including the size and the formats of tuning data, we believe our 3B model gets a better performance mostly because the base model is an instruction-tuned FlanT5, rather than the raw pretrained T5. 
Later we will show that instruction-tuned checkpoint consistently outperforms pretrained checkpoint after specialization (Sec.~\ref{ssec:exp:scaling_curve}), showing the importance of the choice of the base model. 
Also, although not performing well as the teacher model code-davinci-002, our specialized 11B model performance improves to be on par with LaMDA 137B and slightly below PaLM 60B, showing it is indeed possible to make smaller models expert for the particular math reasoning task. 
The price is also very clear: all specialized models suffer from performance drop on BigBench, specifically, they lose all the CoT prompting abilities on BBH, and a large portion of AO prompting performance. 
This observation validates our hypothesis: large models can perform well on a wide range of tasks (here PaLM 60B perform well on both math reasoning and BBH), versus smaller model's ability can be moved from generic tasks (BBH) to a specialized target ability (math reasoning), such that their performance on the target task can still match models that are larger than them, e.g., the average performance on the four math datasets LaMDA 137B 35.9 v.s. specialized FlanT5 11B 40.8.

\begin{table}[t]
  \caption{
  GSM8K validation performance. Instruction-tuned models generally performs better than the raw pretrained checkpoints.
  }
  \label{tab:exp:start_point}
  \centering
  \begin{tabular}{@{}lc|lc@{}}
      \toprule
      \textbf{Before} & \textbf{Acc} & \textbf{After} & \textbf{Acc}  \\ 
      \midrule
      FlanT5 3B & \bf13.5 & Specialized & \bf23.8\\ 
      T5 3B & 0.73 & Specialized & 20.6 \\ \midrule
      FlanT5 760M &  \bf6.9 & Specialized & \bf 21.8  \\ 
      T5 760M &  0.85 & Specialized & 16.2 \\ \midrule
      FlanT5 250M &  \bf 3.0 & Specialized & \bf 15.2  \\ 
      T5 250M &  1.8 & Specialized & 14.2 \\ 
      \bottomrule
  \end{tabular}
\end{table}

\subsection{Scaling Behavior of Smaller Models' CoT Ability}
\label{ssec:exp:scaling_curve}

Now we look the scaling behavoir to smaller models. 
We compare the scaling curve of: 
(1). GPT family small variants (Ada, Babbage, Curie and code-davinci-002);
(2). raw pretrained T5 of different scales and their specialized versions;
(3). the instruction-tuned FlanT5 of different scales and their specialized versions;
The results are shown in Fig.~\ref{fig:exp:scaling} where x-axis denotes the model scale in terms of the number of parameters
and y-axis denotes the validation accuracy on the GSM8K dataset. 

\begin{figure*}[!t]
\small
  \centering
  \includegraphics[width=\linewidth]{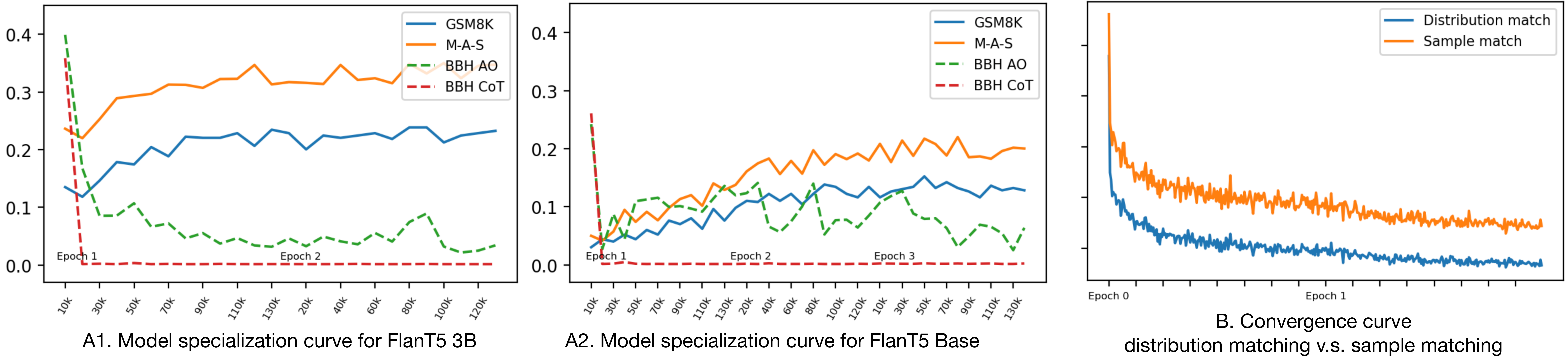}
  \caption{\textbf{A1 and A2}: model specialization curve of FlanT5.
  At the beginning of specialization (e.g., A1 step 10K), the model immediately loses all BBH CoT ability, and a large portion of BBH AO ability. 
  As tuning goes on (e.g., A1 epoch 1), the model's in-distribution performance (GSM8K) and out-of-distribution performance (MultiArith-ASDiv-SVAMP, M-A-S) gradually increases. 
  At the later stage of tuning (e.g., A1 epoch 2), the model's math performance fluctuates and better in-distribution performance does not indicate better out-of-distribution performance. 
  Smaller models need to see the data more times than larger models (A2 has 3 epochs and A1 has 2). 
  \textbf{B}: differences between two distillation approaches. Distribution matching gives faster and lower loss convergence than sampling matching.
  }
  \label{fig:exp:dynamics}
\end{figure*}

\textbf{Smaller models have log-linear, but not flat scaling curve}\quad\quad Initially, in the original CoT paper~\citet{wei2022chain} and the subsequent emergent abilities paper~\citep{wei2022emergent}, CoT prompting is believed to be an emergent property that only large models exhibit.
Smaller model's CoT performance (like smaller GPT variants) was believed to be a flat scaling curve: model performance does not improve with model scale, as is shown in Fig.~\ref{fig:exp:scaling}A left part. 
Later this belief is updated by the FlanT5 paper~\citep{chung2022scaling}, as they show that although the pretrained checkpoint does not have CoT ability, if the model has gone through instruction tuning, smaller models can still exhibit CoT on generic tasks. 
Our work shows that directly trained on CoT data can also lift up the flat scaling curve of the raw T5 checkpoints (Fig.~\ref{fig:exp:scaling}B) to be log-linear.
In Fig.~\ref{fig:exp:scaling}C, we consider specialization for the instruction-tuned FlanT5, and show that specialization significantly lifts up the scaling curve of FlanT5, and both curves are also log-linear. 
All the log-linear curves we observed in Fig.~\ref{fig:exp:scaling} means that the chain-of-thought behavoir of smaller models are not flat, but actually log-linear.
This further indicates that chain-of-thought may not be an emergent ability which
is marked by the flat-then-phase-change curve, but they have the log-linear curve just like large models ~\citep{kaplan2020scaling, hoffmann2022training}. 

\textbf{Instruction-tuned checkpoints perform better than raw pretrained checkpoints}\quad\quad Furthermore, comparing Fig.~\ref{fig:exp:scaling}B and Fig.~\ref{fig:exp:scaling}C, we see that specialized
FlanT5 generally performs better than T5 (though T5 has a larger performance gain). 
The exact validation performance is shown in Table~\ref{tab:exp:start_point}. 
We also believe that, despite there exist multiple confounders, a major reason that our performance in Table~\ref{tab:exp:overall} (FlanT5 11B GSM8K accuracy 27.1) is better than concurrent distillation methods (Magister22 T5 11B, acc. 21.9) is mostly because we use the FlanT5 as our base model versus they use the raw pretrained T5. 
The intuitive explanation is because instruction-tuning elicits the model's full ability while raw pretrained models' ability are not fully released (conceptually see Fig.~\ref{fig:method}A, also see~\citealp{fu2022gptroadmap, chung2022scaling}). 
So for better performance, we recommend using instruction-tuned models in practice. 


\subsection{Specialization Process and Generalization Behaviors}
\label{ssec:exp:dynamics}
Now we consider the specialization process.
Intuitively, during finetuning, the model's ability does not suddenly become the target ability, but will go through a process of moving the models' ability from generic directions to the target. 
We save one checkpoint every 10K instances/ updates,
then evaluate the checkpoints on (1). in-distribution math performance (GSM8K); (2). out-of-distribution math performance (MultiArith, ASDiv, and SVAMP); (3). generic answer-only prompting performance (BBH-AO); (4). generic chain-of-thought prompting performance (BBH-CoT). 
We plot the model's performance across the fine-tuning process in Fig.~\ref{fig:exp:dynamics}.

\textbf{The dynamics of model specilization}. \quad\quad
At the beginning of specialization (Figure A1 at step 10K and Figure A2 at step 20K), 
the model immediately loses all BBH CoT ability (accuracy becomes 0), 
and a large portion of BBH AO ability (accuracy drops from about 0.3 to about 0.1). 
As tuning goes on (A1 epoch 1, A2 epoch 1 and 2), 
the model's in-distribution performance (GSM8K) and out-of-distribution performance (MultiArith-ASDiv-SVAMP, M-A-S) gradually increases, meaning that the model can generalize to three OOD datasets by tuning on GSM8K chain-of-thought data.   
At the later stage of tuning (Figure A1 at epoch 2, and Figure A2 at epoch 3), the model's math performance fluctuates and better in-distribution performance does not indicate better out-of-distribution performance.
The models' BBH-AO performance drops a large portion and the BBH-CoT
performance just die completely. 
Comparing A1 and A2, we also see that smaller models are more data-hungry than larger models~\citep{kaplan2020scaling}:
FlanT5 3B's math performance plateaus at about 90K data points, versus FlanT5 Base's performance continues increase until epoch 3 (each epoch has 130K datapoints). 

\begin{table}[t]
  \caption{
  Model selection method induces tradeoffs between in-distribution and out-of-distribution performance. 
  }
  \label{tab:exp:model_selection}
  \centering
  \begin{tabular}{@{}llcc@{}}
      \toprule
      \textbf{Model} & \textbf{Selection} & \textbf{In-dist} & \textbf{Out-of-dist} \\ \midrule
      FlanT5 3B & GSM8K Dev & 23.8 & 33.2\\
      & M-A-S Dev & 21.2 \blue{-2.6} & 35.0 \green{+1.8} \\ \midrule
      FlanT5 Large & GSM8K Dev & 21.8 & 28.7 \\
      & M-A-S Dev & 19.2 \blue{-2.6} & 30.5 \green{+1.8} \\ \midrule
      FlanT5 Base & GSM8K Dev & 15.2 & 21.7\\
      & M-A-S Dev & 13.2 \blue{-2.0} & 22.0 \green{+0.3} \\ 
      \bottomrule
  \end{tabular}
\end{table}

\textbf{In-distribution and out-of-distribution tradeoffs}  \quad\quad
Because in Fig.~\ref{fig:exp:dynamics} A, both in-distribution and out-of-distribution fluctuates, choosing the best in-distribution checkpoint does not necessarily lead  to the best out-of-distribution checkpoint.
This observation is shown in Table~\ref{tab:exp:model_selection} where if we select the best model based on the GSM8K validation set, it does cannot achieve the best validation performance on the M-A-S OOD setting. 
Yet choosing the best model based on the M-A-S validation performance leads to a smaller performance drop in GSM8K. 
Given this observation, in practice, we would recommend choosing the validation checkpoints according to the specific goal: if the goal is in-distribution generalization, use GSM8K, if the goal is OOD generalization, users may want to use their own validation set (in our case, the M-A-S datasets).


\begin{figure}[!t]
\small
  \centering
  \includegraphics[width=\linewidth]{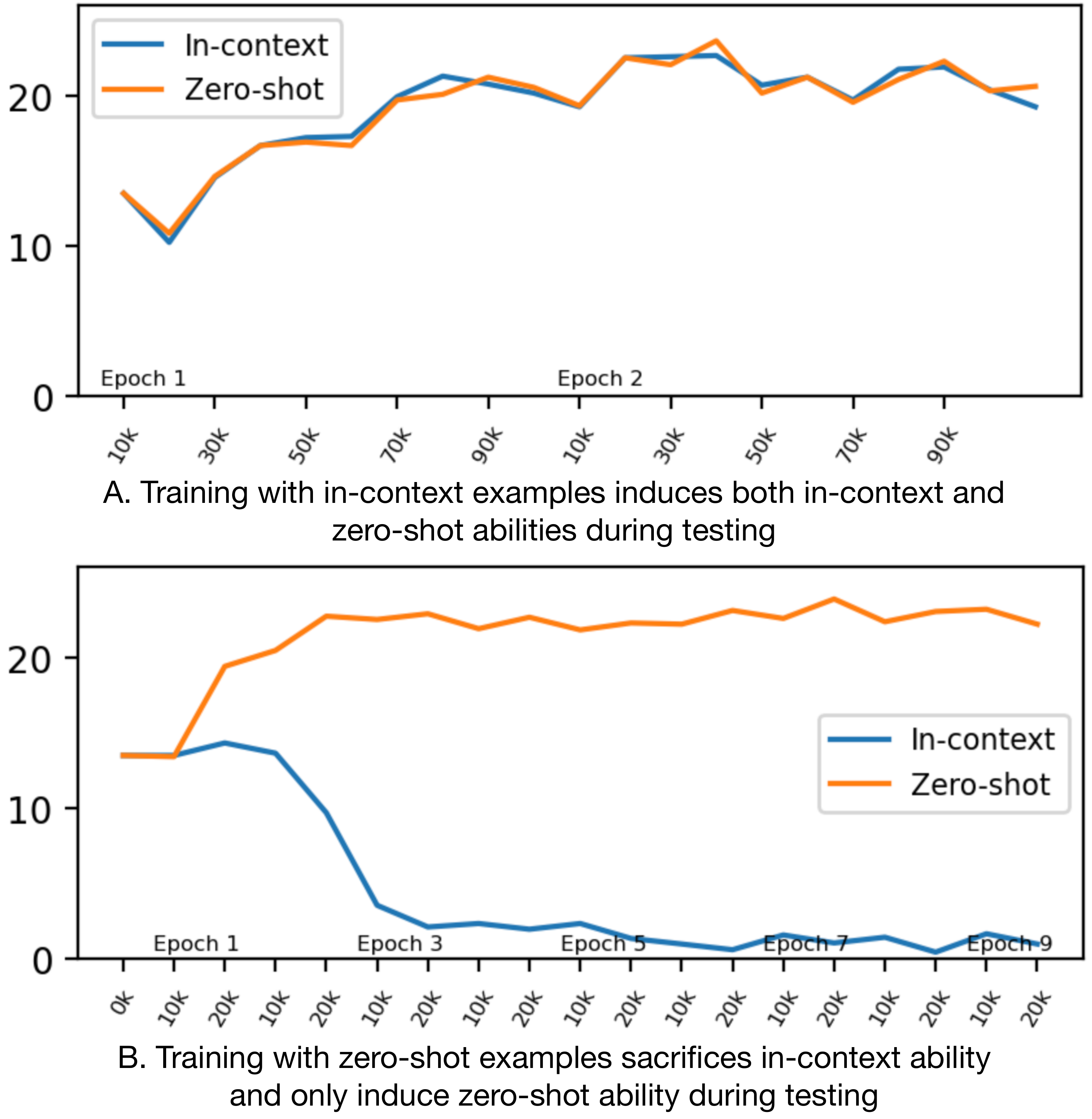}
  \caption{X-axis means tuning datapoints, y-axis means validation accuray on GSM8K. Both figures use FlanT5 3B as the base model.
  \textbf{A}: training with in-context examples automatically give the model zero-shot ability. \textbf{B}: training with zero-shot examples sacrifices in-context ability.
  }
  \label{fig:exp:in_context_zero_shot}
\end{figure}

\subsection{Further Design Choices Analysis}
\label{ssec:exp:analysis}
In this section, we study two more design choices we have discussed before: (1). using distribution matching v.s. sample matching for distillation (recall distillation matching minimizes the KL divergence between FlanT5's per-step autoregressive distribution and GPT's autoregressive distribution, versus sample matching maximizes the likelihood of the reasoning paths generated by GPT);
(2). the influence of data formats, and how in-context/ zero-shot training data induces different behaviors of the specialized model. 

\textbf{Distribution matching gives faster convergence than sample matching}. \quad\quad
Fig.~\ref{fig:exp:dynamics} B shows the training loss of distribution matching v.s. sample matching. 
We show that the model converges faster under distribution matching, and the corresponding loss is lower. 
In terms of validation performance, these two approaches do not differ substantially. 
Yet since distribution matching has a faster convergence, in practice they may still be considered first especially when the model becomes large and tuning becomes expensive. 

\textbf{In-context data preserves zero-shot ability; Zero-shot data loses in-context ability}\quad\quad
This is actually a very interesting observation. 
Specifically, in Fig.~\ref{fig:exp:in_context_zero_shot} A, we tune the model with only in-context data (Format B1 and B2 in Fig~\ref{fig:method}), then test the models in-context learning and zero-shot generalization performance during validation. 
In Fig.~\ref{fig:exp:in_context_zero_shot} B, we tune the model with only zero-shot data (no in-context examples prepended, format B3 and B4 in Fig~\ref{fig:method}), the test if the model can still do in-context learning. 
As is shown in Fig.~\ref{fig:exp:in_context_zero_shot} A, when tuning with in-context data, the model can do both in-context and zero-shot generalization during validation, even the model is not trained with zero-shot data.
In comparison, in Fig.~\ref{fig:exp:in_context_zero_shot} B, when tuning with zero-shot data, the model's zero-shot performance increases, but gradually losses its in-context learning ability. 
This result aligns with the empirical observation on other large models, for example, text-davinci-002 has better zero-shot performance than code-davinci-002, but worse in-context learning performance~\citep{fu2022gptroadmap}. 
This means that the model's ability tradeoff not only happens on math v.s. generic ability, but also happens on zero-shot v.s. in-context learning ability. 
In practice, we would recommend mix the different data formats during tuning (this is why we mix the formats) to maintain a balance between in-context and zero-shot abilities, or adjusting the ratio of different formats according to the specific use case.









\section{Conclusion}
\label{sec:conclusion}
In this work, we study the problem of specializing smaller language models toward multi-step reasoning using chain-of-thought prompting. 
We show that it is indeed possible to concentrate the small models' ability from generic directions to the target math reasoning task. 
After specialization, we show that the model exhibits a log-linear scaling curve where model performance increases smoothly as model scale increases, this is a correction of the previous hypothesis which believes small models have a flat scaling curve that does not increase with model scale. 
We show the importance of using the instruction-tuned checkpoints as the base model because their generalization performance is better than the raw pretrained checkpoints. 
Mutiple tradeoff happens during model specialization, including the loss of BBH performance, the balance between in-distribution and out-of-distribution generalization, and the balance of in-context learning and zero-shot generalization ability. 
We hope our
practice and discoveries can serve as an important
attempt towards specialized smaller models in the
new research paradigm set by LLMs

\nocite{langley00}

\bibliography{references}

\begin{thebibliography}{30}
\providecommand{\natexlab}[1]{#1}
\providecommand{\url}[1]{\texttt{#1}}
\expandafter\ifx\csname urlstyle\endcsname\relax
  \providecommand{\doi}[1]{doi: #1}\else
  \providecommand{\doi}{doi: \begingroup \urlstyle{rm}\Url}\fi

\bibitem[Brown et~al.(2020)Brown, Mann, Ryder, Subbiah, Kaplan, Dhariwal,
  Neelakantan, Shyam, Sastry, Askell, et~al.]{brown2020language}
Brown, T., Mann, B., Ryder, N., Subbiah, M., Kaplan, J.~D., Dhariwal, P.,
  Neelakantan, A., Shyam, P., Sastry, G., Askell, A., et~al.
\newblock Language models are few-shot learners.
\newblock \emph{Advances in neural information processing systems},
  33:\penalty0 1877--1901, 2020.

\bibitem[Chen et~al.(2021)Chen, Tworek, Jun, Yuan, Pinto, Kaplan, Edwards,
  Burda, Joseph, Brockman, et~al.]{chen2021evaluating}
Chen, M., Tworek, J., Jun, H., Yuan, Q., Pinto, H. P. d.~O., Kaplan, J.,
  Edwards, H., Burda, Y., Joseph, N., Brockman, G., et~al.
\newblock Evaluating large language models trained on code.
\newblock \emph{arXiv preprint arXiv:2107.03374}, 2021.

\bibitem[Chowdhery et~al.(2022)Chowdhery, Narang, Devlin, Bosma, Mishra,
  Roberts, Barham, Chung, Sutton, Gehrmann, et~al.]{chowdhery2022palm}
Chowdhery, A., Narang, S., Devlin, J., Bosma, M., Mishra, G., Roberts, A.,
  Barham, P., Chung, H.~W., Sutton, C., Gehrmann, S., et~al.
\newblock Palm: Scaling language modeling with pathways.
\newblock \emph{arXiv preprint arXiv:2204.02311}, 2022.

\bibitem[Chung et~al.(2022)Chung, Hou, Longpre, Zoph, Tay, Fedus, Li, Wang,
  Dehghani, Brahma, et~al.]{chung2022scaling}
Chung, H.~W., Hou, L., Longpre, S., Zoph, B., Tay, Y., Fedus, W., Li, E., Wang,
  X., Dehghani, M., Brahma, S., et~al.
\newblock Scaling instruction-finetuned language models.
\newblock \emph{arXiv preprint arXiv:2210.11416}, 2022.

\bibitem[Cobbe et~al.(2021)Cobbe, Kosaraju, Bavarian, Chen, Jun, Kaiser,
  Plappert, Tworek, Hilton, Nakano, et~al.]{cobbe2021training}
Cobbe, K., Kosaraju, V., Bavarian, M., Chen, M., Jun, H., Kaiser, L., Plappert,
  M., Tworek, J., Hilton, J., Nakano, R., et~al.
\newblock Training verifiers to solve math word problems.
\newblock \emph{arXiv preprint arXiv:2110.14168}, 2021.

\bibitem[Fu et~al.(2022)Fu, Peng, and Khot]{fu2022gptroadmap}
Fu, Y., Peng, H., and Khot, T.
\newblock How does {GPT} obtain its ability? tracing emergent abilities of
  language models to their sources.
\newblock \emph{Yao Fu’s Notion}, Dec 2022.
\newblock URL \url{https://yaofu.notion.site/b9a57ac0fcf74f30a1ab9e3e36fa1dc1}.

\bibitem[Ho et~al.(2022)Ho, Schmid, and Yun]{ho2022large}
Ho, N., Schmid, L., and Yun, S.-Y.
\newblock Large language models are reasoning teachers.
\newblock \emph{arXiv preprint arXiv:2212.10071}, 2022.

\bibitem[Hoffmann et~al.(2022)Hoffmann, Borgeaud, Mensch, Buchatskaya, Cai,
  Rutherford, Casas, Hendricks, Welbl, Clark, et~al.]{hoffmann2022training}
Hoffmann, J., Borgeaud, S., Mensch, A., Buchatskaya, E., Cai, T., Rutherford,
  E., Casas, D. d.~L., Hendricks, L.~A., Welbl, J., Clark, A., et~al.
\newblock Training compute-optimal large language models.
\newblock \emph{arXiv preprint arXiv:2203.15556}, 2022.

\bibitem[Huang et~al.(2022)Huang, Gu, Hou, Wu, Wang, Yu, and
  Han]{huang2022large}
Huang, J., Gu, S.~S., Hou, L., Wu, Y., Wang, X., Yu, H., and Han, J.
\newblock Large language models can self-improve.
\newblock \emph{arXiv preprint arXiv:2210.11610}, 2022.

\bibitem[Iyer et~al.(2022)Iyer, Lin, Pasunuru, Mihaylov, Simig, Yu, Shuster,
  Wang, Liu, Koura, et~al.]{iyer2022opt}
Iyer, S., Lin, X.~V., Pasunuru, R., Mihaylov, T., Simig, D., Yu, P., Shuster,
  K., Wang, T., Liu, Q., Koura, P.~S., et~al.
\newblock Opt-iml: Scaling language model instruction meta learning through the
  lens of generalization.
\newblock \emph{arXiv preprint arXiv:2212.12017}, 2022.

\bibitem[Kaplan et~al.(2020)Kaplan, McCandlish, Henighan, Brown, Chess, Child,
  Gray, Radford, Wu, and Amodei]{kaplan2020scaling}
Kaplan, J., McCandlish, S., Henighan, T., Brown, T.~B., Chess, B., Child, R.,
  Gray, S., Radford, A., Wu, J., and Amodei, D.
\newblock Scaling laws for neural language models.
\newblock \emph{arXiv preprint arXiv:2001.08361}, 2020.

\bibitem[Langley(2000)]{langley00}
Langley, P.
\newblock Crafting papers on machine learning.
\newblock In Langley, P. (ed.), \emph{Proceedings of the 17th International
  Conference on Machine Learning (ICML 2000)}, pp.\  1207--1216, Stanford, CA,
  2000. Morgan Kaufmann.

\bibitem[Li et~al.(2022)Li, Chen, Shen, Chen, Zhang, Li, Wang, Qian, Peng, Mao,
  et~al.]{li2022explanations}
Li, S., Chen, J., Shen, Y., Chen, Z., Zhang, X., Li, Z., Wang, H., Qian, J.,
  Peng, B., Mao, Y., et~al.
\newblock Explanations from large language models make small reasoners better.
\newblock \emph{arXiv preprint arXiv:2210.06726}, 2022.

\bibitem[Liu et~al.(2022)Liu, Lewis, Riedel, and
  Stenetorp]{liu-etal-2022-challenges}
Liu, L., Lewis, P., Riedel, S., and Stenetorp, P.
\newblock Challenges in generalization in open domain question answering.
\newblock In \emph{Findings of the Association for Computational Linguistics:
  NAACL 2022}, pp.\  2014--2029, Seattle, United States, July 2022. Association
  for Computational Linguistics.
\newblock \doi{10.18653/v1/2022.findings-naacl.155}.
\newblock URL \url{https://aclanthology.org/2022.findings-naacl.155}.

\bibitem[Magister et~al.(2022)Magister, Mallinson, Adamek, Malmi, and
  Severyn]{magister2022teaching}
Magister, L.~C., Mallinson, J., Adamek, J., Malmi, E., and Severyn, A.
\newblock Teaching small language models to reason.
\newblock \emph{arXiv preprint arXiv:2212.08410}, 2022.

\bibitem[Min et~al.(2022)Min, Lewis, Zettlemoyer, and
  Hajishirzi]{min-etal-2022-metaicl}
Min, S., Lewis, M., Zettlemoyer, L., and Hajishirzi, H.
\newblock {M}eta{ICL}: Learning to learn in context.
\newblock In \emph{Proceedings of the 2022 Conference of the North American
  Chapter of the Association for Computational Linguistics: Human Language
  Technologies}, pp.\  2791--2809, Seattle, United States, July 2022.
  Association for Computational Linguistics.
\newblock \doi{10.18653/v1/2022.naacl-main.201}.
\newblock URL \url{https://aclanthology.org/2022.naacl-main.201}.

\bibitem[Needleman \& Wunsch(1970)Needleman and Wunsch]{Needleman1970AGM}
Needleman, S.~B. and Wunsch, C.~D.
\newblock A general method applicable to the search for similarities in the
  amino acid sequence of two proteins.
\newblock \emph{Journal of molecular biology}, 48\penalty0 (3):\penalty0
  443--53, 1970.

\bibitem[Ouyang et~al.(2022)Ouyang, Wu, Jiang, Almeida, Wainwright, Mishkin,
  Zhang, Agarwal, Slama, Ray, et~al.]{ouyang2022training}
Ouyang, L., Wu, J., Jiang, X., Almeida, D., Wainwright, C.~L., Mishkin, P.,
  Zhang, C., Agarwal, S., Slama, K., Ray, A., et~al.
\newblock Training language models to follow instructions with human feedback.
\newblock \emph{arXiv preprint arXiv:2203.02155}, 2022.

\bibitem[Raffel et~al.(2020)Raffel, Shazeer, Roberts, Lee, Narang, Matena,
  Zhou, Li, Liu, et~al.]{raffel2020exploring}
Raffel, C., Shazeer, N., Roberts, A., Lee, K., Narang, S., Matena, M., Zhou,
  Y., Li, W., Liu, P.~J., et~al.
\newblock Exploring the limits of transfer learning with a unified text-to-text
  transformer.
\newblock \emph{J. Mach. Learn. Res.}, 21\penalty0 (140):\penalty0 1--67, 2020.

\bibitem[Senin(2008)]{Senin2008DynamicTW}
Senin, P.
\newblock Dynamic time warping algorithm review.
\newblock 2008.

\bibitem[Shridhar et~al.(2022)Shridhar, Stolfo, and
  Sachan]{shridhar2022distilling}
Shridhar, K., Stolfo, A., and Sachan, M.
\newblock Distilling multi-step reasoning capabilities of large language models
  into smaller models via semantic decompositions.
\newblock \emph{arXiv preprint arXiv:2212.00193}, 2022.

\bibitem[Si et~al.(2022)Si, Gan, Yang, Wang, Wang, Boyd-Graber, and
  Wang]{si2022prompting}
Si, C., Gan, Z., Yang, Z., Wang, S., Wang, J., Boyd-Graber, J., and Wang, L.
\newblock Prompting gpt-3 to be reliable.
\newblock \emph{arXiv preprint arXiv:2210.09150}, 2022.

\bibitem[Suzgun et~al.(2022)Suzgun, Scales, Sch{\"a}rli, Gehrmann, Tay, Chung,
  Chowdhery, Le, Chi, Zhou, et~al.]{suzgun2022challenging}
Suzgun, M., Scales, N., Sch{\"a}rli, N., Gehrmann, S., Tay, Y., Chung, H.~W.,
  Chowdhery, A., Le, Q.~V., Chi, E.~H., Zhou, D., et~al.
\newblock Challenging big-bench tasks and whether chain-of-thought can solve
  them.
\newblock \emph{arXiv preprint arXiv:2210.09261}, 2022.

\bibitem[Tan et~al.(2019)Tan, Ren, He, Qin, and Liu]{tan2018multilingual}
Tan, X., Ren, Y., He, D., Qin, T., and Liu, T.-Y.
\newblock Multilingual neural machine translation with knowledge distillation.
\newblock In \emph{International Conference on Learning Representations}, 2019.
\newblock URL \url{https://openreview.net/forum?id=S1gUsoR9YX}.

\bibitem[Tay et~al.(2022)Tay, Dehghani, Tran, Garcia, Bahri, Schuster, Zheng,
  Houlsby, and Metzler]{tay2022unifying}
Tay, Y., Dehghani, M., Tran, V.~Q., Garcia, X., Bahri, D., Schuster, T., Zheng,
  H.~S., Houlsby, N., and Metzler, D.
\newblock Unifying language learning paradigms.
\newblock \emph{arXiv preprint arXiv:2205.05131}, 2022.

\bibitem[Thakkar(2022)]{copilotexplorer}
Thakkar, P.
\newblock Copilot explorer.
\newblock \emph{thakkarparth007.github.io}, 2022.
\newblock URL
  \url{https://thakkarparth007.github.io/copilot-explorer/posts/copilot-internals.html}.

\bibitem[Thoppilan et~al.(2022)Thoppilan, De~Freitas, Hall, Shazeer,
  Kulshreshtha, Cheng, Jin, Bos, Baker, Du, et~al.]{thoppilan2022lamda}
Thoppilan, R., De~Freitas, D., Hall, J., Shazeer, N., Kulshreshtha, A., Cheng,
  H.-T., Jin, A., Bos, T., Baker, L., Du, Y., et~al.
\newblock Lamda: Language models for dialog applications.
\newblock \emph{arXiv preprint arXiv:2201.08239}, 2022.

\bibitem[Wang et~al.(2022)Wang, Wei, Schuurmans, Le, Chi, and
  Zhou]{wang2022self}
Wang, X., Wei, J., Schuurmans, D., Le, Q., Chi, E., and Zhou, D.
\newblock Self-consistency improves chain of thought reasoning in language
  models.
\newblock \emph{arXiv preprint arXiv:2203.11171}, 2022.

\bibitem[Wei et~al.(2022{\natexlab{a}})Wei, Tay, Bommasani, Raffel, Zoph,
  Borgeaud, Yogatama, Bosma, Zhou, Metzler, Chi, Hashimoto, Vinyals, Liang,
  Dean, and Fedus]{wei2022emergent}
Wei, J., Tay, Y., Bommasani, R., Raffel, C., Zoph, B., Borgeaud, S., Yogatama,
  D., Bosma, M., Zhou, D., Metzler, D., Chi, E.~H., Hashimoto, T., Vinyals, O.,
  Liang, P., Dean, J., and Fedus, W.
\newblock Emergent abilities of large language models.
\newblock \emph{Transactions on Machine Learning Research}, 2022{\natexlab{a}}.
\newblock URL \url{https://openreview.net/forum?id=yzkSU5zdwD}.
\newblock Survey Certification.

\bibitem[Wei et~al.(2022{\natexlab{b}})Wei, Wang, Schuurmans, Bosma, Chi, Le,
  and Zhou]{wei2022chain}
Wei, J., Wang, X., Schuurmans, D., Bosma, M., Chi, E., Le, Q., and Zhou, D.
\newblock Chain of thought prompting elicits reasoning in large language
  models.
\newblock \emph{arXiv preprint arXiv:2201.11903}, 2022{\natexlab{b}}.

\end{thebibliography}
\bibliographystyle{icml2023}



\end{document}